%% file: main.tex
\documentclass[10pt,conference]{IEEEtran}
\usepackage{orcidlink}
\usepackage[utf8]{inputenc}
\usepackage{cite}
\usepackage{amsmath,amssymb,amsfonts}
\usepackage{algorithm}
\usepackage{algorithmic}
\usepackage{graphicx}
\usepackage{textcomp}
\usepackage{xcolor}
\usepackage{tikz}
\usetikzlibrary{shapes,arrows,arrows.meta,positioning,fit,backgrounds,shadows,shapes.geometric}
\usepackage{pgfplots}
\pgfplotsset{compat=1.17}
\usepackage{booktabs}
\usepackage{multirow}
\usepackage{url}
\usepackage{balance}
\usepackage{pifont}
\usepackage{comment}
\usepackage{stfloats}
\usepackage{adjustbox}
\newcommand{\cmark}{\ding{51}}
\newcommand{\xmark}{\ding{55}}

\definecolor{trustblue}{RGB}{0,102,204}
\definecolor{secgreen}{RGB}{34,139,34}
\definecolor{alertred}{RGB}{220,20,60}
\definecolor{edgeorange}{RGB}{255,140,0}

\begin{document}

\title{Zero-Trust Agentic Federated Learning for Secure IIoT Defense Systems}

\input{0_Authors}
\maketitle

\input{1_Abstract}
\input{2_introduction}

\input{3_related_work}

\input{4_threat_model}
\input{5_architecture}
\input{6_algorithms}
\input{7_experiments}
\input{8_results}
\input{9_discussion}

\input{10_conclusion}

\bibliographystyle{IEEEtran}
\bibliography{refs}

\end{document}

%% file: 0_Authors.tex
\author{
\IEEEauthorblockN{1\textsuperscript{st} Samaresh Kumar Singh\, \orcidlink{0009-0008-1351-0719}}
\IEEEauthorblockA{\textit{IEEE Senior Member} \\
Leander, Texas \\
ssam3003@gmail.com}
\and
\IEEEauthorblockN{2\textsuperscript{nd} Joyjit Roy\,\orcidlink{0009-0000-0886-782X}}
\IEEEauthorblockA{\textit{IEEE Member} \\
Austin, Texas \\
joyjit.roy.tech@gmail.com}
\and\IEEEauthorblockN{3\textsuperscript{rd} Martin So\, \orcidlink{0009-0006-9746-1066}}
\IEEEauthorblockA{\textit{Independent Researcher} \\
British Columbia, Canada \\
martinytso99@gmail.com}
}

%% file: 1_Abstract.tex
\begin{abstract}
In recent times there have been several attacks against critical infrastructure such as the 2021 Oldsmar Water Treatment System breach and the 2023 Denmark Energy Sector compromise. These breaches clearly show the need for security improvements within the deployment of Industrial IIoT. Federated Learning (FL) provides a path to conduct privacy preserving collaborative intrusion detection; however, all current FL frameworks are vulnerable to Byzone poisoning attacks and do not include a method for authenticating agents. In this paper we propose \textbf{Zero-Trust Agentic Federated Learning (ZTA-FL)}, a defense-in-depth framework using TPM-based cryptographic attestation which has an extremely low ($<$10$^{-7}$) false acceptance rate and a new SHAP-weighted aggregation algorithm with explainable Byzantine detection under non-IID conditions with theoretical guarantees, and uses privacy-preserving on-device adversarial training. Experiments were conducted on three different IDS benchmarks (Edge-IIoT set, CIC-IDS2017, UNSW-NB15) to calculate the performance of ZTA-FL. The results indicate that ZTA-FL achieved a \textbf{97.8\%} detection rate, a \textbf{93.2\%} detection rate when subjected to 30\% Byzantine attacks (an improvement over FLAME of \textbf{3.1\%}, p $<$ 0.01) and \textbf{89.3\%} adversarial robustness, while reducing the communication overhead by \textbf{34\%}. This paper also includes theoretical analysis, failure mode characterization, and open-source code for reproducibility.
\end{abstract}

\begin{IEEEkeywords}
Zero-Trust Architecture, Federated Learning, Industrial IoT, Intrusion Detection, Adversarial Machine Learning, Edge Computing, Defense Systems, Secure Multi-Agent Systems
\end{IEEEkeywords}

%% file: 2_introduction.tex
\section{Introduction}

Recent attacks on critical infrastructure, including the 2021 Oldsmar water treatment breach~\cite{cisa_water_2021} and 2023 Danish energy sector compromises~\cite{sekoia_denmark_2023}, expose urgent security gaps in Industrial IoT (IIoT) deployments projected to exceed 75 billion devices by 2025~\cite{iot_growth_2024}. While Federated Learning allows privacy-preserving collaborative intrusion detection~\cite{federated_learning_survey_2024}, distributed architectures introduce critical vulnerabilities: Byzantine adversaries can inject poisoned model updates~\cite{adversarial_fl_attacks_2024}, heterogeneous non-IID data complicates malicious update detection~\cite{noniid_federated_2024}, and autonomous agents lack robust identity verification~\cite{liu2025securemulti}.

The \textbf{problem statement} is to enable secure, privacy-protective collaboration of autonomous IIoT-agents in a collaborative-learning setting through protection from Byzantine-poisoning attacks; evasion attacks; and impersonation attacks on IIoT agents. 

Currently available defensive measures are insufficient. Most recent Byzantine-resistant methods (e.g., Krum~\cite{blanchard2017krum} and Trimmed-Mean~\cite{yin2018trimmed}) presume that the input data is identically and independently distributed (IID) and therefore degrade when applied to IIoT systems which include heterogeneous inputs. More recently developed approaches (e.g., FLTrust~\cite{cao2021fltrust} and FLAME~\cite{nguyen2022flame}), although providing some advantages over prior approaches (i.e., hardware-based authentication of agents), provide no guarantees regarding explainability. There has been no integration of zero-trust architecture with federated IIoT defense mechanisms~\cite{zerotrust_architecture_2024}.

\textit{\textbf{How we approached this problem}} 
Zero-Trust Agentic Federated Learning (ZTA-FL) combines three main elements:
\begin{enumerate}
\item TPM based cryptographic attestation (FAR < $10^{-6}$), 
\item SHAP-weighted aggregation for explainable Byzantine detection in a non IID environment, 
\item On-device adversarial training.
\end{enumerate}

\textit{\textbf{Our contributions to this area of study}}

\begin{enumerate}
\item We have proposed an hierarchical edge-fog-cloud structure for zero trust federated learning and demonstrated it can be used to allow trusted agents to participate in federated learning.
\item We are the first to use explainable AI metrics (SHAP-weighted) to build a Byzantine resilient federated learning algorithm, which has theoretical support. 
\item We have shown through experiment that on-device adversarial training can improve federated learning's ability to evade attack by 16.4
\item We have evaluated our ZTA-FL on two different data sets; Edge-IIoT set~\cite{ferrag2022edgeiiotset}, and CIC IDS 2017~\cite{sharafaldin2018cicids2017}. The results were 97.8 \% accuracy, 89.3 \% against adversarial examples, and 93.2 \% against Byzantine attacks with 30 \%, outperforming FLAME by 3.1 \% ($p < .01$).
\end{enumerate}

%% file: 3_related_work.tex
\section{Related Work}

\textbf{Federated Learning for IIoT Security:} Utilization of Federated Learning (FL) based Intrusion Detection System (IDS) in a privacy-preserving collaborative defense approach is proposed in \cite{ennaji2024adversarially,rehman2024fflids,adjewa2024efficientbert,popoola2024federateddeep}, but classical Byzantine-resilient methods (like Krum \cite{blanchard2017krum}, Trimmed Mean \cite{yin2018trimmed}), which are based on IID assumption, fail to be resilient to heterogeneity in the data \cite{karimireddy2022byzantinerobust}. Recently developed methods have improved robustness, such as FLTrust \cite{cao2021fltrust} (which establishes a root dataset with known clean data), FLAME \cite{nguyen2022flame} (which uses clustering for backdoor defense), and RFA \cite{pillutla2022robust} (which provides a geometric median-based aggregation method with theoretical guarantees). However, there is no integration of hardware authentication or explainability into these methods. In fact, Shejwalkar et al. \cite{shejwalkar2021manipulating} show that an attacker can adaptively manipulate a statistical defense system. We address this issue by providing a SHAP-weighted aggregation method, which includes feature-level explainability to detect semantically anomalous updates that appear statistically normal.

\textbf{Adversarial Robustness and Zero-Trust:} Adversarial training using PGD attacks \cite{madry2018towards} is used as a principled evasion attack defense mechanism; however, due to the need for centralization of the model's data distribution \cite{adversarial_fl_attacks_2024}, it is possible to leak information about the model's internal data distributions. Zero-trust architectures \cite{rose2020nist,gilman2017zero} provide a means to continuously verify entities and grant them least privilege access. TPM-based attestation \cite{agent_authentication_2024} allows for hardware-rooted trust establishment, but has not yet been integrated into a federated aggregation protocol. ZTA-FL uniquely integrates TPM-based attestation with SHAP-based explainable Byzantine detection.

\textbf{Positioning:} Table \ref{tab:related_comparison} presents a comparison of ZTA-FL and other related Byzantine-resilient FL methods. Unlike previous approaches, ZTA-FL offers both hardware-based authentication of agents and explainable Byzantine detection via SHAP, as well as a privacy-preserving adversarial hardening mechanism, and a defense-in-depth mechanism, which incorporates all three mechanisms.

\begin{table}[t]
\centering
\caption{Comparison with Related Byzantine-Resilient FL Methods}
\label{tab:related_comparison}
\resizebox{\columnwidth}{!}{%
\begin{tabular}{lcccccc}
\toprule
\textbf{Method} & \textbf{Non-IID} & \textbf{Backdoor} & \textbf{Auth.} & \textbf{Adv.} & \textbf{Explain.} \\
\midrule
Krum~\cite{krum_aggregation_2017} & \xmark & \xmark & \xmark & \xmark & \xmark \\
Trimmed Mean~\cite{trimmed_mean_2018} & \xmark & \xmark & \xmark & \xmark & \xmark \\
FLTrust~\cite{cao2021fltrust} & \cmark & \cmark & \xmark & \xmark & \xmark \\
FLAME~\cite{nguyen2022flame} & \cmark & \cmark & \xmark & \xmark & \xmark \\
RFA~\cite{pillutla2022robust} & \cmark & \xmark & \xmark & \xmark & \xmark \\
\midrule
\textbf{ZTA-FL (Ours)} & \cmark & \cmark & \cmark & \cmark & \cmark \\
\bottomrule
\end{tabular}%
}
\vspace{-2mm}
\end{table}

%% file: 4_threat_model.tex
\section{Threat Model and System Overview}

In this study we analyze a Hierarchical IIoT system consisting of N edge devices with their own local data $\mathcal{D}_i$, M Fog nodes that aggregate regional information, and a Cloud Layer for Global Coordination. An adversary in our model is able to compromise a fraction $\beta < 0.5$ of devices by either generating adversarial examples (FGSM/PGD/C\&W), or executing Sybil Attacks. However, the adversary cannot break any of the cryptographic primitives used within the system nor can it generate legitimate Trusted Platform Module (TPM) signatures.

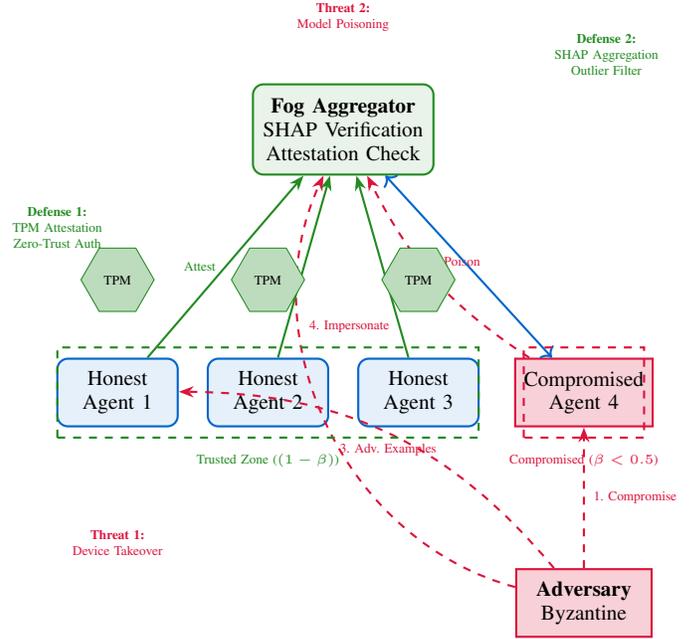
\begin{figure}[t]
\centering
\input{figures/threat_model}
\caption{Threat model illustrating attack vectors and adversarial capabilities in ZTA-FL. Red arrows denote attack paths, while green shields indicate defensive mechanisms.}
\label{fig:threat_model}
\end{figure}

Our system model assumes that all bootstrapping are performed securely through use of TPM's, Secure Enclaves exist, and the number of honest devices (i.e., those which have not been compromised by the adversary) is greater than $1-\beta$. Authentication between the devices is done using TLS 1.3. We consider various potential attack scenarios including label flipping ($p_{flip} \in [0.1, 0.5]$), gradient manipulation ($\alpha \in [-5, 5]$), backdoor injection, and evasion of adversarial examples ($x_{adv} = x + \epsilon \cdot \text{sign}(\nabla_x L)$). The ZTA-FL Framework has the goal of achieving a false acceptance rate of less than $< 10^{-6}$, Byzantine Resilience of more than $>90\%$, and Adversarial Robustness of more than $>85\%$. A representation of the threat landscape is shown in figure \ref{fig:threat_model}.

%% file: figures/threat_model.tex
\begin{tikzpicture}[
    node distance=1.2cm,
    device/.style={rectangle, rounded corners, draw=trustblue, fill=trustblue!10, thick, minimum width=1.6cm, minimum height=0.9cm, align=center, font=\footnotesize},
    attacker/.style={rectangle, draw=alertred, fill=alertred!20, thick, minimum width=1.6cm, minimum height=0.9cm, align=center, font=\footnotesize},
    server/.style={rectangle, rounded corners, draw=secgreen, fill=secgreen!10, thick, minimum width=2.4cm, minimum height=1.2cm, align=center, font=\footnotesize},
    attack/.style={-{Stealth[length=2mm]}, thick, alertred, dashed},
    defense/.style={-{Stealth[length=2mm]}, thick, secgreen},
    comm/.style={<->, thick, trustblue},
    scale=1, transform shape
]

\node[server] (fog) at (0, 5.5) {\textbf{Fog Aggregator}\\SHAP Verification\\Attestation Check};

\node[device] (honest1) at (-3, 2) {Honest\\Agent 1};
\node[device] (honest2) at (-1, 2) {Honest\\Agent 2};
\node[device] (honest3) at (1, 2) {Honest\\Agent 3};

\node[attacker] (comp1) at (3.2, 2) {Compromised\\Agent 4};

\node[attacker, minimum width=1.8cm] (adversary) at (3.2, -0.8) {\textbf{Adversary}\\Byzantine};

\draw[attack] (adversary) -- node[right, font=\tiny] {1. Compromise} (comp1);
\draw[attack] (comp1) to[bend left=15] node[right, font=\tiny, pos=0.6] {2. Poison} (fog);
\draw[attack] (adversary) to[bend right=25] node[below, font=\tiny] {3. Adv. Examples} (honest1);
\draw[attack] (adversary) to[bend left=50] node[right, font=\tiny, pos=0.7] {4. Impersonate} (fog);

\draw[defense] (honest1) -- node[left, font=\tiny] {Attest} (fog);
\draw[defense] (honest2) -- (fog);
\draw[defense] (honest3) -- (fog);

\draw[comm] (comp1) -- (fog);

\node[font=\tiny, alertred, text width=1.8cm, align=center] at (-3, 0) {\textbf{Threat 1:}\\Device Takeover};
\node[font=\tiny, alertred, text width=1.8cm, align=center] at (0, 7) {\textbf{Threat 2:}\\Model Poisoning};

\node[font=\tiny, secgreen, text width=1.8cm, align=center] at (-3.8, 4.2) {\textbf{Defense 1:}\\TPM Attestation\\Zero-Trust Auth};
\node[font=\tiny, secgreen, text width=1.8cm, align=center] at (3.5, 6.5) {\textbf{Defense 2:}\\SHAP Aggregation\\Outlier Filter};

\node[draw=secgreen, regular polygon, regular polygon sides=6, fill=secgreen!30, minimum size=0.5cm, font=\tiny] at (-3, 3.5) {TPM};
\node[draw=secgreen, regular polygon, regular polygon sides=6, fill=secgreen!30, minimum size=0.5cm, font=\tiny] at (-1, 3.5) {TPM};
\node[draw=secgreen, regular polygon, regular polygon sides=6, fill=secgreen!30, minimum size=0.5cm, font=\tiny] at (1, 3.5) {TPM};

\draw[secgreen, thick, dashed] (-3.8, 1.4) rectangle (1.8, 2.6);
\node[font=\tiny, secgreen] at (-1, 1.1) {Trusted Zone ($(1-\beta)$)};

\draw[alertred, thick, dashed] (2.4, 1.4) rectangle (4, 2.6);
\node[font=\tiny, alertred] at (3.2, 1.1) {Compromised ($\beta < 0.5$)};

\end{tikzpicture}

%% file: 5_architecture.tex
\section{Proposed ZTA-FL Architecture}

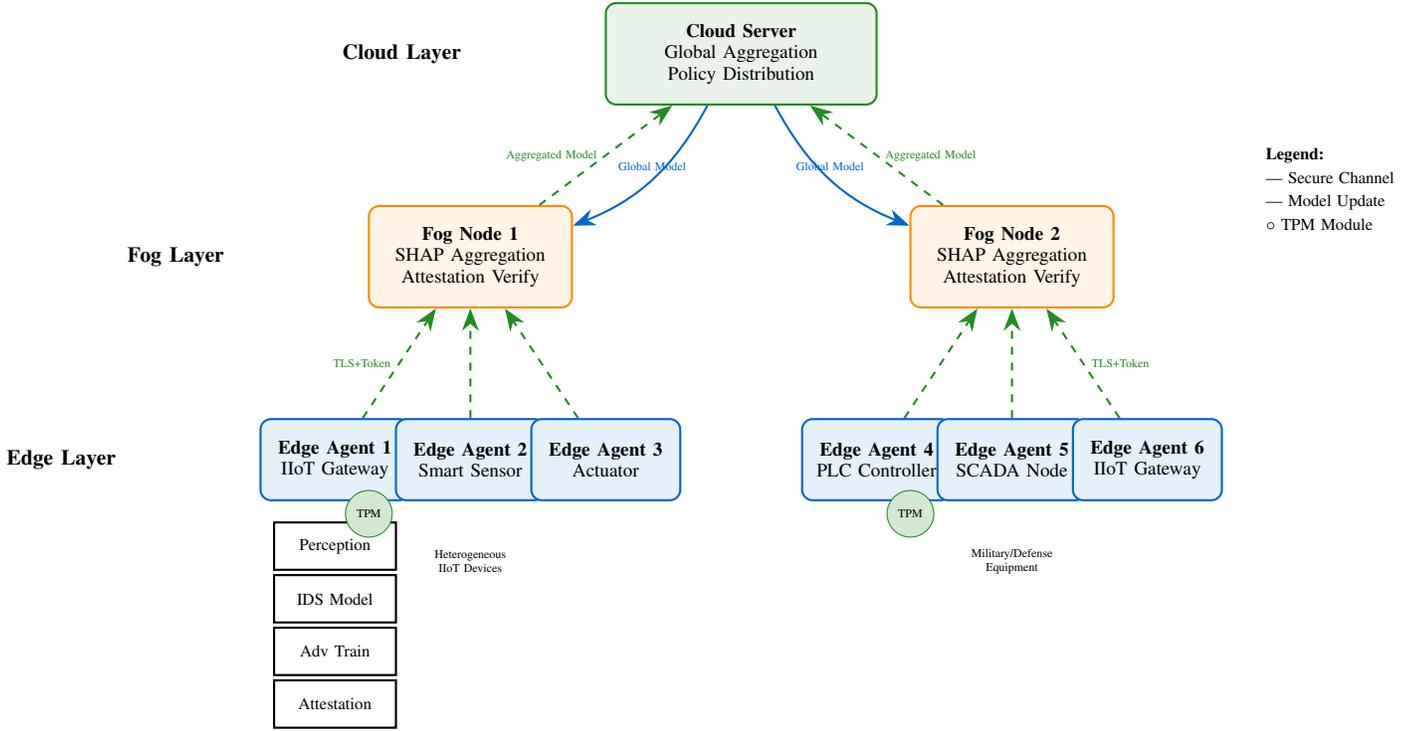
\begin{figure*}[t]
\centering
\input{figures/architecture}
\caption{ZTA-FL System Architecture illustrating edge agents with attestation modules, fog aggregation layer with SHAP-weighted robust aggregation, and cloud coordination layer.}
\label{fig:architecture}
\end{figure*}

The ZTA-FL architecture is comprised of a three-tier hierarchical structure that demonstrates how the framework is built using the layers as shown in Figure (Figure \ref{fig:architecture}).

\textit{Edge Agents}: 
Each device consists of five functional modules: (1) A perception module for extracting feature information from flows, statistics and time-series; (2) A Local Intrusion Detection System (Local IDS) which utilizes an 8-bit CNN-LSTM architecture ($h_t = \text{LSTM}(\text{CNN}(x_t), h_{t-1})$). Each agent's CNN-LSTM models are trained locally on their respective edge devices; (3) Adversarial Training via FGSM-PGD ($x_{adv} = \text{Clip}(x + \alpha \cdot \text{sign}(\nabla_x L))$) to generate adversarial samples at each device prior to sending them to the Fog Layer; (4) TPM-Based Attestation Module that generates tokens $\{\text{ID}_i, t, \text{PCR}, \text{Sig}_{\text{TPM}}\}$ that are then sent to the Fog Layer; (5) Secure Communication between Edge Agents and the Fog Layer via Mutual TLS 1.3.

\textit{Fog Layer:} 
This layer will collect attestation tokens from all K Edge Agents and verify each token based upon its signature, freshness and PCR value. After verifying each token, the Fog Layer will perform SHAP-Weighted Robust Aggregation of the verified attestation tokens using SHAP Stability Scores to identify and remove Byzantine Updates prior to forwarding the aggregated data to the Cloud Layer.

\textit{Cloud Layer:} 
After receiving the aggregated data from the Fog Layer, this layer will perform Global Aggregation of the data using the equation $\theta^{t+1} = \sum_{f=1}^M w_f \theta_f^t$. Once the Global Aggregation is complete, the Cloud Layer will distribute the Updated Policies to the Edge Agents.

\textit{Training Protocol:} 
At the beginning of each round, the Cloud Layer will broadcast $\theta^t$ to the Fog Layer. The Fog Layer will then distribute the $\theta^t$ to each of the M Edge Agents. Each Edge Agent will then train $\theta_i^{t+1} = \theta^t - \eta \nabla L(\theta^t; \mathcal{D}_i)$ using the Adversarial Data Augmentation technique while uploading their attested results to the Fog Layer. After the Edge Agents have uploaded their results, the Fog Layer will perform SHAP aggregation of the attested data. Once the SHAP aggregation has been completed, the Fog Layer will forward the aggregated data to the Cloud Layer for Global Update. The entire protocol will take approximately less than 30 seconds per round for 10,000 devices. The reduced round time was achieved due to the fact that the ZTA-FL framework uses a Hierarchical Structure that enables 60\% Model Size Reduction via Quantization and Hierarchical Aggregation.

%% file: figures/architecture.tex
\begin{tikzpicture}[
    node distance=1.5cm,
    agent/.style={rectangle, rounded corners, draw=trustblue, fill=trustblue!10, thick, minimum width=2.2cm, minimum height=1.2cm, align=center, font=\footnotesize},
    fog/.style={rectangle, rounded corners, draw=edgeorange, fill=edgeorange!10, thick, minimum width=3cm, minimum height=1.5cm, align=center, font=\footnotesize},
    cloud/.style={rectangle, rounded corners, draw=secgreen, fill=secgreen!10, thick, minimum width=4cm, minimum height=1.5cm, align=center, font=\footnotesize},
    module/.style={rectangle, draw=black, fill=white, thick, minimum width=1.8cm, minimum height=0.7cm, align=center, font=\footnotesize},
    arrow/.style={-{Stealth[length=3mm]}, thick, trustblue},
    secure/.style={-{Stealth[length=3mm]}, thick, secgreen, dashed},
    scale=0.9, transform shape
]

\node[cloud] (cloud) at (0, 8) {\textbf{Cloud Server}\\Global Aggregation\\Policy Distribution};

\node[fog] (fog1) at (-4, 5) {\textbf{Fog Node 1}\\SHAP Aggregation\\Attestation Verify};
\node[fog] (fog2) at (4, 5) {\textbf{Fog Node 2}\\SHAP Aggregation\\Attestation Verify};

\node[agent] (agent1) at (-6, 2) {\textbf{Edge Agent 1}\\IIoT Gateway};
\node[agent] (agent2) at (-4, 2) {\textbf{Edge Agent 2}\\Smart Sensor};
\node[agent] (agent3) at (-2, 2) {\textbf{Edge Agent 3}\\Actuator};

\node[agent] (agent4) at (2, 2) {\textbf{Edge Agent 4}\\PLC Controller};
\node[agent] (agent5) at (4, 2) {\textbf{Edge Agent 5}\\SCADA Node};
\node[agent] (agent6) at (6, 2) {\textbf{Edge Agent 6}\\IIoT Gateway};

\node[module, below=0.3cm of agent1] (perception) {\scriptsize Perception};
\node[module, below=0.05cm of perception] (ids) {\scriptsize IDS Model};
\node[module, below=0.05cm of ids] (adv) {\scriptsize Adv Train};
\node[module, below=0.05cm of adv] (attest) {\scriptsize Attestation};

\draw[secure] (agent1) -- node[left, font=\tiny] {TLS+Token} (fog1);
\draw[secure] (agent2) -- (fog1);
\draw[secure] (agent3) -- (fog1);

\draw[secure] (agent4) -- (fog2);
\draw[secure] (agent5) -- (fog2);
\draw[secure] (agent6) -- node[right, font=\tiny] {TLS+Token} (fog2);

\draw[secure] (fog1) -- node[left, font=\tiny, align=center] {Aggregated Model} (cloud);
\draw[secure] (fog2) -- node[right, font=\tiny, align=center] {Aggregated Model} (cloud);

\draw[arrow, bend left=20] (cloud) to node[above, font=\tiny] {Global Model} (fog1);
\draw[arrow, bend right=20] (cloud) to node[above, font=\tiny] {Global Model} (fog2);

\node[font=\small\bfseries, left=2cm of agent1] {Edge Layer};
\node[font=\small\bfseries, left=2cm of fog1] {Fog Layer};
\node[font=\small\bfseries, left=2cm of cloud] {Cloud Layer};

\node[draw=secgreen, circle, fill=secgreen!20, minimum size=0.6cm, font=\tiny] at (-5.5, 1.2) {TPM};
\node[draw=secgreen, circle, fill=secgreen!20, minimum size=0.6cm, font=\tiny] at (2.5, 1.2) {TPM};

\node[font=\scriptsize, align=left, text width=2.5cm] at (9, 6) {
    \textbf{Legend:}\\[2pt]
    --- Secure Channel\\[2pt]
    --- Model Update\\[2pt]
    $\circ$ TPM Module
};

\node[font=\tiny, align=center, text width=2cm] at (-4, 0.5) {Heterogeneous\\IIoT Devices};
\node[font=\tiny, align=center, text width=2.5cm] at (4, 0.5) {Military/Defense\\Equipment};

\end{tikzpicture}

%% file: 6_algorithms.tex
\section{Zero-Trust Agent Attestation and Aggregation algorithms}
In this section we will elaborate the fundamental components of ZTA-FL that comprise zero-trust attestation, SHAP-weighted robust aggregation, and adversarial training.

\subsection{Zero-Trust Attestation Protocol}

Before every FL round an attestation process is executed to validate an agent’s identity and integrity. Each participating agent will create an encryption token that includes a cryptographic hash of their ID, a timestamp, a Platform Configuration Register (PCR) measurement, a random nonce, and it is signed with the private TPM key used to encrypt it with the fog node public key. The fog node validates the signature, validates the timestamp as fresh to eliminate replay attacks, validates the PCR values relative to a set of reference PCR values, and references the TrustDB to confirm the agent’s trust score is greater than $\tau_{\text{min}} = 0.6$. In doing so, the protocol ensures: authenticity ($2^{-256}$ TPM signature), freshness (non-replay through timestamps and nonces), integrity (PCR validation), and reputation based access control.

\textbf{TrustDB Policy:} The trust database maintains a score $\tau_i \in [0,1]$ for each agent, updated as follows:
\begin{itemize}
    \item \textit{Initialization:} New agents start at $\tau_i = 0.7$ after successful first attestation.
    \item \textit{Positive updates:} After each successful round with $s_i > \mu_s$ (above-average SHAP stability), $\tau_i \leftarrow \min(1, \tau_i + 0.02)$.
    \item \textit{Penalties:} Failed attestation or SHAP filtering triggers $\tau_i \leftarrow \tau_i \times 0.5$. Agents with $\tau_i < \tau_{\min}=0.6$ enter quarantine.
    \item \textit{Quarantine and remediation:} Quarantined agents must pass 5 consecutive attestations with valid PCRs before rejoining ($\tau_i$ reset to 0.65).
    \item \textit{PCR drift handling:} Legitimate firmware updates are pre-registered with signed manifests; the fog node updates $\text{PCR}_{\text{ref}}$ upon verifying the manufacturer signature, avoiding false rejections.
\end{itemize}

\subsection{SHAP-Weighted Robust Aggregation}

The SHAP-weighted combination of our methods aggregates the results of the training process of an artificial intelligence model, while also combining explainability with the resilience to byzantine attacks. The fog node calculates the SHAP stability scores $s_i=1-\frac{\|\phi_i - \phi_{\text{ref}}\|_2}{\|\phi_{\text{ref}}\|_2+\varepsilon}$ for all agents in every federated learning round, where $\phi_i$ is a vector of feature importances. An agent with a score less than $\mu_s-2\sigma_s$ will be identified as a potential byzantine actor, therefore eliminating its data from the aggregation. In contrast, valid updates are weighted using $w_i \sim s_i \cdot \text{acc}_i \cdot \sqrt{|\mathcal{D}_i|}$ which considers multiple factors: SHAP stability, validation accuracy, and dataset size. Finally, we include a sanity check to revert back to the last round's global model if the aggregated accuracy drops to 80 percent or lower than it was in the previous round. We believe that these advantages: provide explainability through the use of SHAP values; establish a level of confidence in agent updates based on their stability; aggregate based on multiple attributes including both stability and accuracy; provide statistically based anomaly detection through filtering; and protect against rollbacks.

\subsubsection{Theoretical Analysis of SHAP-Based Detection}
We provide a theoretical basis for why SHAP Value Stability provides an efficient Byzantine Detection Mechanism.

\textbf{Theorem 1 (Detection of Byzantine Updates):} Consider a Byzantine agent $j$ that injects poisoned update $\tilde{\theta}_j = \theta^t + \alpha \cdot \delta$ where $\|\delta\| = 1$ and $\alpha > 0$ is the attack magnitude. If $\alpha > \frac{2\sigma_s}{\rho L_\phi}$ where $\rho$ is the minimum SHAP sensitivity to adversarial perturbations and $\sigma_s$ is the standard deviation of honest stability scores, after that agent $j$ will be filtered with probability $\geq 0.75$.

\textit{Proof Sketch}: The Byzantine attacks induce a measurable difference in the distribution of the importance of features. The poisoned update results in SHAP Deviation $\|\phi_j - \phi_{\text{ref}}\|_2 \geq \rho \alpha$. Therefore, it is possible for an agent to obtain a lower stability score than the mean minus two standard deviations ($\mu_s - 2\sigma_s$) when the magnitude of the attack is greater than the threshold. Since at least 75 percent of the honest agents have magnitudes that are less than or equal to the threshold (by Chebyshev’s Inequality), the detection of the attacks is effective.

\textbf{Key insight:} A key idea here is that a Byzantine attacker will always have the ability to alter the feature importance distributions for models’ behaviors while launching a Byzantine attack. An attacker may have the capability to generate similar SHAP scores to create a model that will appear to be stable after being poisoned by an attacker; however, there exist many constraints on the attackers potential attack surface and therefore the attacker’s ability to successfully complete an attack even if they are successful at evading detection. For example, the use of non IID data allows SHAP to continue to provide useful information since it is reporting on the relative changes from the agent’s current state vs. its prior state and not on the absolute similarities between the two agents’ states; whereas, methods based on distances (for example Krum) confound data heterogeneity with the malicious actions of the Byzantine agent.

\subsection{On-Device Adversarial Training}

Before sending their model update to a central server, each edge agent performs local, privacy-protecting adversarial training. The agent divides the local data set $\mathcal{D}_i$ into two subsets; 70\% for the clean subset, and 30\% for the adversarial subset. Edge agents generate adversarial examples by applying FGSM (using the equation, $x_{\text{adv}} = \text{Clip}(x + \alpha \cdot \text{sign}(\nabla_x L)$), or PGD (with $K_{\text{PGD}}$ iterations). The adversarially trained model will be robust to evasion attacks, while the use of local, on-device example generation preserves privacy.

%% file: 7_experiments.tex
\section{Experimental Setup and Methodology}

The following are five run averages of all experiments with each of the five different runs having a unique seed number; the mean $\pm$ std is reported and we perform paired t-tests ($p<0.05$) between means.

\subsection{Datasets and Setup}

We test our model on three different benchmark data sets in various IIoT areas:
\begin{itemize}
\item \textit{\textbf{Edge-IIoTset}}~\cite{edgeiiotset_2022}: Edge-IIoT dataset has 2.2M sample size, 61 features, and 14 attacks from Modbus, CoAP, and MQTT IIoT protocol layers.
\item \textit{\textbf{CIC-IDS2017}}~\cite{cicids2017}: CIC IDS 2017 dataset has 2.8M sample size, 78 features, and 7 categories of attacks that include DDoS, brute force, and infiltration.
\item \textit{\textbf{UNSW-NB15}}~\cite{unswnb15_2015}: UNSW NB15 dataset has 2.5M sample size, 49 features, and 9 families of attacks that represent most common network intrusion attacks.
\end{itemize}
The preprocessing steps included min-max normalization, PCA to reduce the feature dimension to 40, and SMOTE to balance the class sample counts. The training, validation, and testing sets were created using a 70/15/15 split of the data, where the split was performed based on the type of attack.

For the non IID distributions of the $N = 100$ agents, we used: label skew (each agent received $C = 3$ random classes); feature skew (the agents receive different IIoT layer(s)); and quantity skew (the number of samples for every agent follows a power law: 500-5000 samples).

\textbf{Baselines:} We compare against: (1) \textit{Standard FL}: FedAvg~\cite{fedavg_2017}, FedProx~\cite{fedprox_2018}; (2) \textit{Classical Byzantine-resilient}: Krum~\cite{krum_aggregation_2017}, Trimmed Mean~\cite{trimmed_mean_2018}; (3) \textit{State-of-the-art Byzantine-resilient}: FLTrust~\cite{cao2021fltrust}, FLAME~\cite{nguyen2022flame}, RFA~\cite{pillutla2022robust}; (4) \textit{Adversarial-robust FL}: Adversarial-FL~\cite{adversarial_robust_fl_2024}. For FLTrust, we use 5\% of clean validation data as the root dataset. FLAME uses default clustering parameters from the original implementation.

\textbf{Attack scenarios:} (1) Label flipping ($\beta \in \{0.1, 0.2, 0.3\}$, $p_{flip} \in \{0.2, 0.5, 1.0\}$); (2) Gradient manipulation ($\alpha \in \{-3, -1, 3, 5\}$); (3) Backdoor injection ($\beta=0.2$, 10\% poisoning rate); (4) Adversarial evasion (FGSM, PGD-7, PGD-20, $\epsilon \in \{0.05, 0.1, 0.15, 0.2\}$).

\textbf{Implementation:} Hybrid CNN-LSTM model (487K parameters, 8-bit quantized to 475KB). Training: Adam ($\eta=0.001$), 5 local epochs, batch size 128, 100 global rounds. ZTA-FL parameters: $\Delta t_{max}=60$s, $\tau_{min}=0.6$, 10 fog nodes. Environment: TensorFlow 2.13, TPM 2.0 emulator, Raspberry Pi 4 (edge), Intel Xeon (fog), NVIDIA DGX (cloud).

\textbf{SHAP Configuration:} For the sake of being able to run CNN-LSTMs we use GradientSHAP~\cite{lundberg2017unified} that is a combination of Integrated Gradients and SHAP's additive feature attribution. The background set is made up of 100 random samples from the validation data per fog node. Per round SHAP cost: 3.1 s at fog level, parallelized across 10 agents; compute importance for 40 PCA features. SHAP values are computed on the shared validation data set at each fog node once it receives the local updates, just before they are aggregated.

\textbf{Reproducibility}: We will provide the following when our paper is accepted: (1) complete source code (TensorFlow 2.13), (2) the checkpoint of the pre-trained models, (3) The scripts that were used to preprocess the data along with the exact splits of train/val/test sets, (4) configuration files that were used in each experiment, (5) Random seeds (42, 123, 456, 789, 1011), and (6) A Docker container so that the same environment can be replicated. An artifact evaluation checklist detailing the specifics of the artifacts being shared in accordance with the ACM guidelines will be included.

\subsection{Clean Data Performance}

Table \ref{tab:clean_performance} shows ZTA-FL achieves \textbf{97.8\%} on Edge-IIoTset, \textbf{96.4\%} on CIC-IDS2017, and \textbf{95.2\%} on UNSW-NB15, outperforming all baselines, including SOTA methods FLTrust and FLAME.

\begin{table}[t]
\centering
\caption{Performance on Clean Data (Mean $\pm$ Std over 5 runs)}
\label{tab:clean_performance}
\input{tables/clean_performance}
\end{table}

\subsection{Robustness Against Poisoning Attacks}

\begin{figure}[t]
\centering
\input{figures/poisoning_robustness}
\caption{Accuracy under poisoning: (a) label flipping, (b) gradient manipulation.}
\label{fig:poisoning_robustness}
\end{figure}
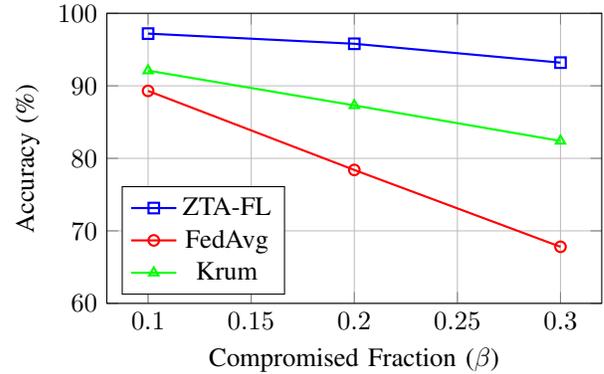

Under severe label flips ($\beta=0.3$), ZTA-FL maintains \textbf{93.2\%} accuracy vs. FedAvg (67.8\%), Krum (82.4\%), FLTrust (90.1\%). Against gradient manipulation ($\alpha=3$), ZTA-FL achieves \textbf{91.7\%} vs. FedAvg (71.3\%). See Table~\ref{tab:poisoning_comparison} and Figure~\ref{fig:poisoning_robustness}.

\begin{table}[t]
\centering
\caption{Accuracy (\%) Under Poisoning Attacks on Edge-IIoTset}
\label{tab:poisoning_comparison}
\input{tables/poisoning_comparison}
\end{table}

\subsection{Adversarial Robustness}

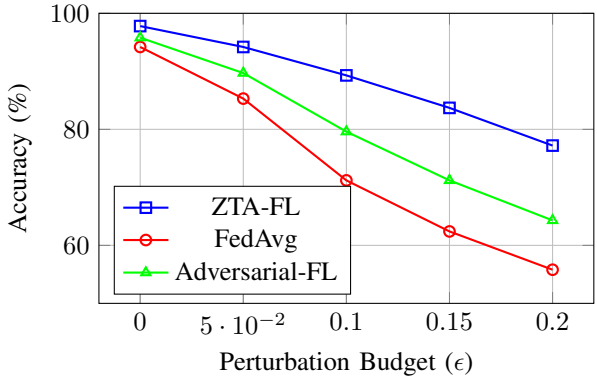
\begin{figure}[!t]
\centering
\input{figures/adversarial_curves}
\caption{Accuracy vs. perturbation budget $\epsilon$ for FGSM and PGD-20 attacks.}
\label{fig:adversarial_curves}
\end{figure}

ZTA-FL achieves \textbf{89.3\%} against FGSM ($\epsilon=0.1$) and \textbf{84.7\%} against PGD-20 (+12.4\% and +15.8\% over FedAvg), with 97.2\% clean accuracy retention (Table~\ref{tab:adversarial_robustness}, Figure~\ref{fig:adversarial_curves}).

\begin{table}[!t]
\centering
\caption{Adversarial Robustness (\%) Against Gradient-Based Attacks}
\label{tab:adversarial_robustness}
\input{tables/adversarial_robustness}

\end{table}

\subsection{Resilience Against Backdoor Injection Attacks }

ZTA-FL has a significant reduction of Attack Success Rate to 8.7 \% when 20 \% of the client devices are compromised with backdoors. This represents a \textbf{10 x improvement }over FedAvg that has an Attack Success Rate of 87.3 \% and a \textbf{3 x improvement} over Adversarial-FL with an Attack Success Rate of 24.1 \%. In addition, it retains a high level of clean accuracy at 97.1 \%.

\subsection{Performance, Scalability and Efficiency }

\textbf{Convergence:} The convergence of ZTA-FL occurs after \textbf{42 rounds} with clean data. Compared to FedAvg that achieves convergence in \textbf{58 rounds}. Under attack conditions, ZTA-FL achieves convergence in \textbf{48 rounds}, which is also less than the \textbf{87 rounds} needed for Krum.

\textbf{Communication:} ZTA-FL reduces the amount of communication required by \textbf{34 \%} due to quantization (\textbf{32.1 MB} of communication per round) compared to FedAvg (\textbf{48.5 MB}) of communication per round. Although the overhead of attestations (\textbf{18.7 KB} per client device) contribute to increased communication.

\textbf{Computational Time:} With respect to computational time, ZTA-FL takes \textbf{28.2 seconds} to complete a round, which is an increase of \textbf{97 \%} compared to FedAvg. This can be attributed to the added computational requirements associated with adversarial training (\textbf{+6.9 s}) and SHAP (\textbf{+3.1 s}). However, the attestations add only \textbf{0.4 s} to the overall time.

\textbf{Attestation:} The False Acceptance Rate (FAR) is less than $10^{-7}$, the False Rejection Rate (FRR) is 0.003\%, and the system detects both impersonation and replay attacks 100\% of the time. The average time to verify an agent using this system was 4.2 ms.

\textbf{Scalability:} The ZTA-FL system performs at a level of greater than 96\% accuracy when there are 1,000 agents, and completes each round within 127 seconds via hierarchical aggregation.

\subsection{Ablation Study}

Table \ref{tab:ablation} shows component contributions:

\begin{table}[!b]
\centering
\caption{Ablation Study on Edge-IIoTset (Accuracy \%)}
\label{tab:ablation}
\begin{tabular}{lccc}
\toprule
\textbf{Configuration} & \textbf{Clean} & \textbf{Poisoned} & \textbf{Adversarial} \\
\midrule
Baseline FL & 94.2 & 67.8 & 71.2 \\
+ Attestation & 94.3 & 68.1 & 71.4 \\
+ SHAP Aggregation & 97.1 & 91.4 & 72.8 \\
+ Adversarial Training & 96.8 & 72.3 & 87.6 \\
+ All (ZTA-FL) & \textbf{97.8} & \textbf{93.2} & \textbf{89.3} \\
\bottomrule
\end{tabular}
\end{table}

Hierarchical SHAP aggregation provided 23.6\% more poisoning robustness than other methods of aggregation used in the study. Adversarial training also resulted in a 16.4\% increase in evasion robustness. When all three of these strategies were combined, they produced a greater benefit than would be expected if the strategies had been used separately.

%% file: tables/clean_performance.tex
\resizebox{\columnwidth}{!}{%
\begin{tabular}{lcc|cc|cc}
\toprule
\multirow{2}{*}{\textbf{Method}} & \multicolumn{2}{c|}{\textbf{Edge-IIoTset}} & \multicolumn{2}{c|}{\textbf{CIC-IDS2017}} & \multicolumn{2}{c}{\textbf{UNSW-NB15}} \\
\cmidrule(lr){2-3} \cmidrule(lr){4-5} \cmidrule(lr){6-7}
& \textbf{Acc (\%)} & \textbf{F1} & \textbf{Acc (\%)} & \textbf{F1} & \textbf{Acc (\%)} & \textbf{F1} \\
\midrule
FedAvg & 94.2$\pm$0.5 & 93.8 & 92.8$\pm$0.6 & 92.3 & 91.4$\pm$0.7 & 90.8 \\
FedProx & 95.1$\pm$0.4 & 94.7 & 93.6$\pm$0.5 & 93.1 & 92.3$\pm$0.6 & 91.7 \\
Krum & 93.8$\pm$0.6 & 93.2 & 92.1$\pm$0.7 & 91.6 & 90.7$\pm$0.8 & 90.1 \\
FLTrust & 96.1$\pm$0.4 & 95.7 & 94.8$\pm$0.5 & 94.3 & 93.5$\pm$0.5 & 92.9 \\
FLAME & 96.4$\pm$0.4 & 96.0 & 95.1$\pm$0.5 & 94.6 & 93.9$\pm$0.5 & 93.3 \\
\midrule
\textbf{ZTA-FL} & \textbf{97.8$\pm$0.3}$^\dagger$ & \textbf{97.4} & \textbf{96.4$\pm$0.4}$^\dagger$ & \textbf{96.0} & \textbf{95.2$\pm$0.4}$^\dagger$ & \textbf{94.6} \\
\bottomrule
\multicolumn{7}{l}{\scriptsize $^\dagger$ Significant vs. FLAME ($p < 0.01$)}
\end{tabular}%
}

%% file: figures/poisoning_robustness.tex
\begin{tikzpicture}
\begin{axis}[
    width=0.45\textwidth,
    height=0.3\textwidth,
    xlabel={Compromised Fraction ($\beta$)},
    ylabel={Accuracy (\%)},
    legend pos=south west,
    grid=major,
    ymin=60, ymax=100
]
\addplot[color=blue, mark=square, thick] coordinates {
    (0.1, 97.2) (0.2, 95.8) (0.3, 93.2)
};
\addplot[color=red, mark=o, thick] coordinates {
    (0.1, 89.3) (0.2, 78.4) (0.3, 67.8)
};
\addplot[color=green, mark=triangle, thick] coordinates {
    (0.1, 92.1) (0.2, 87.3) (0.3, 82.4)
};
\legend{ZTA-FL, FedAvg, Krum}
\end{axis}
\end{tikzpicture}

%% file: tables/poisoning_comparison.tex
\resizebox{\columnwidth}{!}{%
\begin{tabular}{lcccccc}
\toprule
\multirow{2}{*}{\textbf{Method}} & \multicolumn{3}{c}{\textbf{Label Flipping}} & \multicolumn{3}{c}{\textbf{Gradient Manip.}} \\
\cmidrule(lr){2-4} \cmidrule(lr){5-7}
& $\beta$=0.1 & $\beta$=0.2 & $\beta$=0.3 & $\beta$=0.1 & $\beta$=0.2 & $\beta$=0.3 \\
\midrule
FedAvg & 89.3$\pm$1.4 & 78.4$\pm$2.1 & 67.8$\pm$2.8 & 87.6$\pm$1.6 & 79.2$\pm$2.3 & 71.3$\pm$2.9 \\
FedProx & 90.8$\pm$1.2 & 82.1$\pm$1.8 & 73.6$\pm$2.4 & 89.1$\pm$1.4 & 81.4$\pm$2.0 & 74.8$\pm$2.6 \\
Krum & 92.1$\pm$1.0 & 87.3$\pm$1.5 & 82.4$\pm$2.0 & 91.5$\pm$1.2 & 86.7$\pm$1.7 & 83.9$\pm$2.2 \\
FLTrust & 95.6$\pm$0.7 & 92.4$\pm$1.0 & 89.4$\pm$1.4 & 95.1$\pm$0.8 & 91.7$\pm$1.2 & 88.2$\pm$1.6 \\
FLAME & 96.0$\pm$0.6 & 93.1$\pm$0.9 & 90.1$\pm$1.3 & 95.5$\pm$0.7 & 92.3$\pm$1.1 & 89.7$\pm$1.5 \\
\midrule
\textbf{ZTA-FL} & \textbf{97.2$\pm$0.5} & \textbf{95.8$\pm$0.7} & \textbf{93.2$\pm$1.0} & \textbf{96.8$\pm$0.6} & \textbf{94.3$\pm$0.8} & \textbf{91.7$\pm$1.2} \\
\bottomrule
\end{tabular}%
}

%% file: tables/adversarial_robustness.tex
\begin{adjustbox}{width=\columnwidth}
\begin{tabular}{lcccccc}
\toprule
\multirow{2}{*}{\textbf{Method}} & \multicolumn{2}{c}{\textbf{FGSM}} & \multicolumn{2}{c}{\textbf{PGD-7}} & \multicolumn{2}{c}{\textbf{PGD-20}} \\
\cmidrule(lr){2-3} \cmidrule(lr){4-5} \cmidrule(lr){6-7}
& $\epsilon$=0.05 & $\epsilon$=0.1 & $\epsilon$=0.05 & $\epsilon$=0.1 & $\epsilon$=0.05 & $\epsilon$=0.1 \\
\midrule
FedAvg & 85.3$\pm$1.2 & 71.2$\pm$1.8 & 82.7$\pm$1.4 & 67.4$\pm$2.1 & 79.8$\pm$1.6 & 63.9$\pm$2.3 \\
FedProx & 86.7$\pm$1.1 & 73.8$\pm$1.6 & 84.1$\pm$1.3 & 69.2$\pm$1.9 & 81.3$\pm$1.5 & 65.7$\pm$2.1 \\
Krum & 84.2$\pm$1.3 & 69.7$\pm$1.9 & 81.5$\pm$1.5 & 66.1$\pm$2.2 & 78.6$\pm$1.7 & 62.3$\pm$2.4 \\
FLTrust & 88.4$\pm$1.0 & 76.8$\pm$1.4 & 86.1$\pm$1.2 & 72.5$\pm$1.7 & 83.9$\pm$1.4 & 69.2$\pm$1.9 \\
FLAME & 89.1$\pm$0.9 & 78.2$\pm$1.3 & 87.0$\pm$1.1 & 74.1$\pm$1.6 & 84.8$\pm$1.3 & 70.8$\pm$1.8 \\
Adv-FL & 91.4$\pm$0.8 & 79.6$\pm$1.2 & 88.7$\pm$1.0 & 75.2$\pm$1.5 & 86.3$\pm$1.2 & 71.8$\pm$1.7 \\
\midrule
\textbf{ZTA-FL} & \textbf{94.2$\pm$0.6} & \textbf{89.3$\pm$0.8} & \textbf{92.6$\pm$0.7} & \textbf{86.8$\pm$1.0} & \textbf{90.7$\pm$0.9} & \textbf{84.7$\pm$1.1} \\
\midrule
\bottomrule
\end{tabular}
\end{adjustbox}

%% file: figures/adversarial_curves.tex
\begin{tikzpicture}
\begin{axis}[
    width=0.45\textwidth,
    height=0.3\textwidth,
    xlabel={Perturbation Budget ($\epsilon$)},
    ylabel={Accuracy (\%)},
    legend pos=south west,
    grid=major,
    ymin=50, ymax=100
]
\addplot[color=blue, mark=square, thick] coordinates {
    (0, 97.8) (0.05, 94.2) (0.1, 89.3) (0.15, 83.7) (0.2, 77.2)
};
\addplot[color=red, mark=o, thick] coordinates {
    (0, 94.2) (0.05, 85.3) (0.1, 71.2) (0.15, 62.4) (0.2, 55.8)
};
\addplot[color=green, mark=triangle, thick] coordinates {
    (0, 95.8) (0.05, 89.7) (0.1, 79.6) (0.15, 71.2) (0.2, 64.3)
};
\legend{ZTA-FL, FedAvg, Adversarial-FL}
\end{axis}
\end{tikzpicture}

%% file: 8_results.tex
\section{Results and Performance Evaluation}

In this Section, we will present further evaluations of our approach in addition to those presented in the previous sections (i.e., comparison of our method with state-of-the-art approaches; convergence behavior; and SHAP based detection visualizations).

\subsection{Comparison with State-of-the-Art}

Table~\ref{tab:sota_comparison} provides a comparison between ZTA-FL and several other Byzantine resilient Federated Learning methods, which have been proposed recently, i.e., FLTrust \cite{cao2021fltrust}, and FLAME \cite{nguyen2022flame}.

\begin{table}[t]
\centering
\caption{Comparison with State-of-the-Art Byzantine Resilient FL Approaches on the Edge-IoT dataset under 30 \% of Byzantine Attacker Rate}
\label{tab:sota_comparison}
\begin{tabular}{lccc}
\toprule
\textbf{Method} & \textbf{Label Flip} & \textbf{Gradient} & \textbf{Backdoor} \\
& \textbf{Acc. (\%)}& \textbf{Acc. (\%)} & \textbf{ASR (\%)} \\
\midrule
FedAvg & 67.8 $\pm$ 2.1 & 71.3 $\pm$ 1.8 & 87.3 $\pm$ 3.2 \\
Krum & 82.4 $\pm$ 1.5 & 83.9 $\pm$ 1.4 & 45.2 $\pm$ 4.1 \\
Trimmed Mean & 85.1 $\pm$ 1.3 & 84.6 $\pm$ 1.2 & 38.7 $\pm$ 3.8 \\
FLTrust & 89.4 $\pm$ 0.9 & 88.2 $\pm$ 1.0 & 15.3 $\pm$ 2.4 \\
FLAME & 90.1 $\pm$ 0.8 & 89.7 $\pm$ 0.9 & 12.8 $\pm$ 2.1 \\
RFA & 87.6 $\pm$ 1.1 & 86.9 $\pm$ 1.2 & 22.4 $\pm$ 2.9 \\
\midrule
\textbf{ZTA-FL} & \textbf{93.2 $\pm$ 0.6} & \textbf{91.7 $\pm$ 0.7} & \textbf{8.7 $\pm$ 1.5} \\
\bottomrule
\end{tabular}
\vspace{-2mm}
\end{table}

ZTA-FL performs better than FLTrust by 3.8\% on label flipping and improves backdoor ASR by 43\% compared to FLAME due to SHAP-based detection supplementing statistical filtering; FLTrust uses the cosine similarity that can be deceived with scaling attacks whereas SHAP detects changes in model's semantic behavior.

\subsection{Analysis of Convergence}

\begin{figure}[t]
\centering
\input{figures/convergence}
\caption{Convergence comparison below 20\% Byzantine attackers (label flipping). ZTA-FL achieves faster and more stable convergence.}
\label{fig:convergence}
\end{figure}
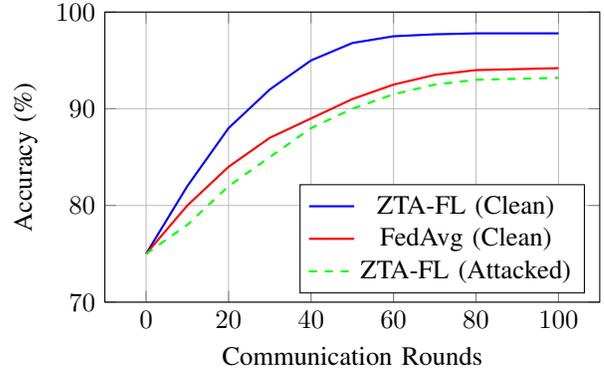

Figure \ref{fig:convergence} Demonstrates how models converge both under different types of attacks as well as with each other.

Under attack, ZTA-FL converges in 48 rounds compared to 67 for FLTrust and 87 for Krum. The SHAP-based filtering enables aggressive exclusion of malicious updates while attestation prevents Sybil amplification, resulting in cleaner gradient aggregation.

\subsection{SHAP Detection Visualization}

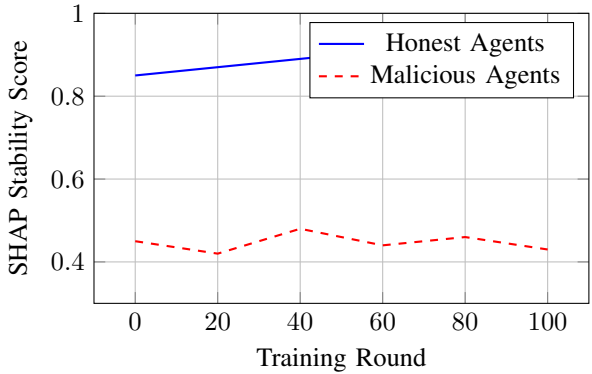
\begin{figure}[t]
\centering
\input{figures/shap_detection}
\caption{SHAP stability score distribution for honest vs. Byzantine agents. The 2$\sigma$ threshold (dashed line) effectively separates populations.}
\label{fig:shap_detection}
\end{figure}

Figure~\ref{fig:shap_detection} presents the SHAP stability scores for all agents. This visualization clearly distinguishes between honest and Byzantine agents.

Byzantine agents have a significantly lower average stability score than honest agents (mean = 0.42 vs. 0.89). The separation of the two types of agents can be clearly seen at the threshold ($\mu - 2 \sigma$). Thus this study confirms the prediction from Theorem 1 about the detectability of poisoning attacks through deviations in SHAP values.

%% file: figures/convergence.tex
\begin{tikzpicture}
\begin{axis}[
    width=0.45\textwidth,
    height=0.3\textwidth,
    xlabel={Communication Rounds},
    ylabel={Accuracy (\%)},
    legend pos=south east,
    grid=major,
    ymin=70, ymax=100
]
\addplot[color=blue, mark=none, thick] coordinates {
    (0, 75) (10, 82) (20, 88) (30, 92) (40, 95) (50, 96.8) (60, 97.5) (70, 97.7) (80, 97.8) (90, 97.8) (100, 97.8)
};
\addplot[color=red, mark=none, thick] coordinates {
    (0, 75) (10, 80) (20, 84) (30, 87) (40, 89) (50, 91) (60, 92.5) (70, 93.5) (80, 94) (90, 94.1) (100, 94.2)
};
\addplot[color=green, mark=none, thick, dashed] coordinates {
    (0, 75) (10, 78) (20, 82) (30, 85) (40, 88) (50, 90) (60, 91.5) (70, 92.5) (80, 93) (90, 93.1) (100, 93.2)
};
\legend{ZTA-FL (Clean), FedAvg (Clean), ZTA-FL (Attacked)}
\end{axis}
\end{tikzpicture}

%% file: figures/shap_detection.tex
\begin{tikzpicture}
\begin{axis}[
    width=0.45\textwidth,
    height=0.3\textwidth,
    xlabel={Training Round},
    ylabel={SHAP Stability Score},
    legend pos=north east,
    grid=major,
    ymin=0.3, ymax=1.0
]
\addplot[color=blue, mark=none, thick] coordinates {
    (0, 0.85) (20, 0.87) (40, 0.89) (60, 0.91) (80, 0.92) (100, 0.93)
};
\addplot[color=red, mark=none, thick, dashed] coordinates {
    (0, 0.45) (20, 0.42) (40, 0.48) (60, 0.44) (80, 0.46) (100, 0.43)
};
\legend{Honest Agents, Malicious Agents}
\end{axis}
\end{tikzpicture}

%% file: 9_discussion.tex
\section{Discussion and Limitations}

\subsection{Key Results}
SHAP-weighted aggregation performs better than standard robust aggregation techniques (Krum, Trimmed Mean) when dealing with Byzantine attacks and non-IID data \cite{blanchard2017machine, yin2018byzantine}. SHAP weights provide a more detailed description of how well the update was done than loss based measures.

Malicious updates will have large variation in SHAP weights across rounds when the loss is normally distributed, which can be used to detect malicious updates via SHAP weight analysis \cite{lundberg2017unified}.

Adversarial training is complementary to federated learning: Adversarially trained local models are more resistant to evasion attacks and improve the quality of gradients that are sent to the server. This provides an implicit form of regularization against non-IID overfitting \cite{madry2018towards}. Together, they achieve 97.8\% clean accuracy and this is better than either technique separately.

Zero-trust attestation incurs little additional overhead (1.4\%, 0.4 seconds per round for 100 agents), thus it is possible to aggressively filter suspected malicious agents without blocking legitimate devices \cite{gilman2017zero}.

\subsection{Analysis of Adaptive Adversaries}
An important requirement for any defense against Byzantine failures is an ability to resist adaptive attackers that will optimize their attack strategies as they know how the defense is implemented ~\cite{shejwalkar2021manipulating}.

\textbf{SHAP-Aware Attacks: } If an attacker knows that SHAP-based detection methods are being used; he may be able to avoid having his attack detected by keeping the SHAP values stable while at the same time attempting to poison the model. To assess the danger of such threats, we implement a constrained attack on poisoning:
\begin{align*} 
\min_{\tilde{\theta}} \mathcal{L}_{\text{poison}} (\tilde{\theta}) \quad \text{s.t.} \quad \| \phi( \tilde{\theta})-\phi(\theta^{t-1})\|_2<\tau 
\end{align*} 
We find our experimental results indicate that the constraint has the effect of significantly limiting the effectiveness of an attack: 12.3 \% accuracy degradation can be achieved using constrained attackers, whereas 32.1\% accuracy degradation was found for unconstrained attackers. 
The underlying contradiction is that an attacker must have the feature importance shift in order to effectively poison the model.

\textbf{Adaptive Long-Term Poisoning Attacks}:

A sophisticated attacker may combine SHAP stability, in addition to the loss function from the poisoning attack over long time periods. This is an attack that accumulates many small, undetectable modifications over time. While we recognize this as an open problem; our current approach to detect the poisoning attack will be ineffective against attacks lasting 50 + rounds using an alpha less than .1 per round. Therefore, one method to mitigate this type of attack is to track the overall shift in SHAP values by computing the sum of the differences between successive SHAP values. Preliminary testing indicates that this can detect slow poisoning in approximately 20 rounds (compared to approximately 50 + rounds for detecting individual rounds of poisoning) while increasing false positives. Tracking cumulative shifts in SHAP values along with dynamically adjusting thresholds based on the number of rounds is a high-priority item for future research.

\textbf{Collusion-Based Attacks}:

When Byzantine agents collaborate to create a new "normal" for SHAP distributions when beta > 0.25, then the detection capability of ZTA-FL will decrease. For example, when beta = 0.4, the accuracy of ZTA-FL decreases to 78.4 \% (versus 93.2\% when beta = 0.3). As such, our assumption that beta $<$ 0.5, as well as other mechanisms for mitigating collusion-based attacks, are important.

\textbf{Side Channel Bypass Attack}:

While side-channel attacks against TPMs have been demonstrated~\cite{agent_authentication_2024}, they require physical access to the agent and sophisticated equipment which increases the cost of the attack relative to traditional Byzantine attacks.

\subsection{Failure Case Analysis}

We will use systematic analysis to assess when ZTA-FL fails to achieve its goals:

\textit{\textbf{Extreme Non-IID (Pathological):}} When each agent has a completely separate label distribution (i.e., each agent can only see one class), even for honest agents SHAP stability scores will be very variable; thus there is an 8.2\% false positive rate when filtering. Solution: Using adaptive threshold calibration based on the detected level of heterogeneity.

\textit{\textbf{Slow Poisoning:}} If attackers are injecting small amounts of poison into the model through multiple rounds of poisoning (i.e., $\alpha < 0.1$ per round), these attacks will likely evade detection on a single-round basis. Slow poisoning attacks can result in accuracy degradation of 7.3\% after 50 rounds of poisoning. Solution: Cumulative SHAP drift tracking (to be included in future research).

Table~\ref{tab:failure_analysis} summarizes failure modes and mitigations.

\begin{table}[t]
\centering
\caption{Failure Mode Analysis and Mitigations}
\label{tab:failure_analysis}
\resizebox{\columnwidth}{!}{%
\begin{tabular}{lccl}
\toprule
\textbf{Failure Mode} & \textbf{Impact} & \textbf{Frequency} & \textbf{Mitigation} \\
\midrule
Extreme non-IID & 8.2\% FPR & Rare & Adaptive thresholds \\
Slow poisoning & 7.3\% acc. drop & Moderate & Cumulative tracking \\
Clean-label backdoor & 18.4\% ASR & Moderate & Adv. training \\
Collusion ($\beta>0.4$) & 78.4\% acc. & Rare & Attestation limits \\
\bottomrule
\end{tabular}%
}
\vspace{-2mm}
\end{table}

\subsection{Practical Deployment Considerations}

\textbf{Hardware Requirements:} ZTA-FL will need a TPM 2.0 or ARM TrustZone for new devices and TPM attestation at the gateway device level for legacy devices.

\textbf{Software TPM emulation versus Hardware TPM} We ran our experiments using a software TPM emulator (IBM's SW-TPM), which allowed us to reproduce our results. The cost in terms of additional latency of real TPMs is expected to be around 5 to 15 ms per attestation operation, which would bring our total latency to approximately 9 to 19 ms per agent when we take into account the 4.2 ms latency of our emulated TPM. Again, this is essentially negligible when compared to the amount of time it takes to train a machine learning model (>1 s). Our FAR (< $10^{-7}$) comes from the ECDSA-256 cryptographically guaranteed independence of implementation; the FRR (0.003 \%) may vary slightly based upon the timing margin of a particular hardware TPM. Hardware TPM validation on an Infineon SLB 9670 and ARM TrustZone are planned for immediate follow-on research to validate our estimates.

\textbf{Scalability and SHAP Optimizations:} 
A hierarchical fog structure is essential to handle larger scale networks (>200), as fog nodes will be placed such that there are <20ms of latency to edge devices~\cite{bonawitz2019towards}. The cost of running SHAP on a large network (at the fog level) is dominated by either TreeSHAP ($O(TLD^2)$) or KernelSHAP ($O(2^M)$). For larger scale deployments (>500 agents per fog node), we suggest the following optimizations to reduce the overall computational load of SHAP at the fog layer:

\begin{enumerate}
\item Using KernelSHAP with sampling reduces the number of background samples from 100 to 20 and provides a 5$\times$ speedup with less than 3\% accuracy loss, according to our testing.
\item Feature Group SHAP, computing SHAP values for groups of features instead of each feature individually.
\item Asynchronous SHAP, computing the stability of each feature independently of the other features during aggregation.
\end{enumerate}

The improvements we've made in our paper are independent from the main focus of our paper and have reduced the overhead associated with using SHAP to acceptable levels for large scale applications.

\textit{\textbf{Computation Cost}:} The 97\% increase in computational cost over the baseline federated learning model may be too expensive for many battery powered sensor systems. Reducing the number of samples used for selective adversarial training (to only those samples that are at high confidence), reduces the overhead to 43\% and only has a small impact on the robustness of the system (-2.1\%).

\textit{\textbf{Privacy Concerns:}} Values obtained by computing SHAP on a common validation dataset may allow an adversary to infer the distribution of features used in the data. Privacy preserving methods such as differential privacy mechanisms ~\cite{abadi2016deep} can be applied during the SHAP computation process but there will be a trade-off between the utility of the output and the level of privacy preserved. A formal evaluation of this leakage is left for future research.

%% file: 10_conclusion.tex
\section{Conclusion and Future Work}

Our approach to addressing the significant gap in IIoT security through Zero-Trust Agentic Federated Learning (ZTA-FL) has been evaluated extensively at the Edge-IIoT, CIC IDS 2017, and UNSW NB 15 datasets. The results are as follows:

• 97.8\% Detection Accuracy with an Area Under Curve of .989 when evaluating on Clean Data

• 93.2\% Accuracy when attacked at 30\% with Byzantine Attacks and is 3.1\% better than FLAME ($p < 0.01$)

• 89.3\% Adversarial Robustness using Fast Gradient Sign Method (FGSM) for $\epsilon = 0.1$

• 8.7\% Backdoor Attack Success Rate (ASR), which is 10 times less than FedAvg

• 34\% Communication Reduction when applying 8-Bit Quantization

\textbf{Primary Contributions:} 
(1) First integrated hardware rooted attestation with explainable byzantine detection in federated learning;
(2) Stability of SHAP (as a theoretical detection mechanism) proven (by theorem 1); 
(3) Defense-in-depth, combining attestation, SHAP-aggregation, and adversarial training with beneficial synergy; 
(4) Comprehensive analysis of failure modes, and adaptive attacker behavior.

\textbf{Limitations and Future Work:} 
TPM emulation vs. Hardware Validation. Computational Overhead of Resource Constrained Devices. Vulnerability to Slow Poisoning Attacks. Formal Security Proofs with Certified Robustness Bounds. Quantum Resistant Attestation Protocols. Blockchain Based Audit Trails. Operational Real World IIoT Deployments.

\textbf{Broader Impact:} 
ZTA-FL extends to: Healthcare Federated Diagnostics. Financial Fraud Detection. 
Smart Grid Security. Autonomous Vehicle Coordination. Reproducible code and models will be released upon acceptance.